\documentclass{article}
\usepackage{authblk}
\usepackage{amsfonts}

\usepackage{algorithm}
\usepackage{algpseudocode}
\usepackage[autostyle=true]{csquotes}

\usepackage{hyperref}
\usepackage[utf8]{inputenc}
\usepackage[textwidth=14cm]{geometry}
\usepackage{float, graphicx, amsmath, xcolor, caption, subcaption, hyperref}
\usepackage[sorting=none, style=numeric-comp]{biblatex}
\bibliography{references} 

\title{PredProp: Bidirectional Stochastic Optimization \\ with Precision Weighted Predictive Coding}
\author[1*]{Andr\'{e} Ofner}
\author[1]{Sebastian Stober}
\affil{Otto-von-Guericke University, Magdeburg, Germany \protect\\ ofner@ovgu.de}

\begin{document}

\maketitle

\begin{abstract}
We present PredProp, a method for optimization of weights and states in predictive coding networks (PCNs) based on the precision of propagated errors and neural activity. PredProp jointly addresses inference and learning via stochastic gradient descent and adaptively weights parameter updates by approximate curvature. Due to the relation between propagated error covariance and the Fisher information matrix, PredProp implements approximate Natural Gradient Descent. We demonstrate PredProp's effectiveness in the context of dense decoder networks and simple image benchmark datasets. We found that PredProp performs favorably over Adam, a widely used adaptive learning rate optimizer in the tested configurations. Furthermore, available optimization methods for weight parameters benefit from using PredProp's error precision during inference. Since hierarchical predictive coding layers are optimised individually using local errors, the required precisions factorize over hierarchical layers. Extending beyond classical PCNs with a single set of decoder layers per hierarchical layer, we also generalize PredProp to deep neural networks in each PCN layer by additionally factorizing over the weights in each PCN layer. 

\end{abstract}

\section{Introduction}
In the context of machine learning, neural networks are often trained by updating a set of parameters $p$ in order to optimize an objective function $L(p)$. Stochastic Gradient Descent (SGD) is a widely used iterative method that optimizes the objective function by updating the parameters with randomly sampled batches from a dataset using the negative gradient of the cost function \cite{robbins1951stochastic}. Under the assumption that derivatives for the parameters can be computed, this results in simple update rules for the parameters $p_t$ at discrete steps $t$ in the form of:

\begin{equation}
p_{t+1} = p_{t}+\Delta p_{t} = p_{t} - \eta g_{t}
\end{equation}

where $g_{t}=\frac{\partial L\left(p_{t}\right)}{\partial p_{t}}$ is the gradient of the parameters at step $t$ and $\eta$ is a learning rate that modulates the size of the update.

Training of neural networks typically focuses on a global optimization of weight parameters and inference is usually restricted to a single step, by computing the network's prediction given a particular input. Recently, (gradient based) optimisation of generative models, consisting of a hierarchy of state parameters and multiple, locally optimised weights has seen increased attention in neuroscience and machine learning research \cite{friston2010generalised, parr2019neuronal, millidge2021predictivereview, csenoz2018online, scellier2017equilibrium}. Our focus here is predictive coding (PC) on static inputs, a general approach to addressing state inference and parameter learning in complex generative models \cite{friston2009predictive, millidge2021predictivereview, ofner2022generalized, zahid2023curvature}. In predictive coding networks (PCNs), each hierarchical layer is optimised independently using local prediction and prediction error signals. Because a PCN's parameter space is factorized into multiple smaller parts, this lets us consider optimisation methods, such as (approximate) natural gradient descent (NGD) \cite{amari1998natural}. NGD can be tricky to implement and is computationally expensive in the context of deep neural networks without such factorized parameter sets \cite{martens2015optimizing, lin2021tractable}. Next to weights learning, state inference in PCNs can be cast as an iterative gradient descent procedure as well, allowing us to employ advanced optimisation strategies during state updates.

\section{Predictive coding networks}
Predictive coding is a theory that originates from cognitive neuroscience and aims at explaining brain function \cite{rao1999predictive, friston2003learning, friston2006free, bastos2012canonical, millidge2021predictivereview}. It offers a description of neural operations required for maintaining a hierarchical generative model using a relatively simple algorithmic motif based on prediction error minimisation and bidirectional processing \cite{rao1999predictive, friston2003learning, friston2009predictive, marino2022predictive}. PCNs optimise neural activity and, on a slower timescale, neural plasticity in order to minimize the prediction error between an observed stimulus and its prediction. 

In more detail, PCNs infer weights $\theta$ and (cause) state parameters $v$ of a generative model $p(u ; \theta)=\int p(u \mid v ; \theta) p(v ; \theta) dv$ from data $u$. 

PCNs aim at encoding the conditional distribution of the cause states given observed data, by inverting the generative model:

\begin{equation}
p(v \mid u ; \theta)=\frac{p(u \mid v ; \theta) p(v ; \theta)}{p(u ; \theta)}
\end{equation}

To do so, PCNs optimise an approximate density $q(v)$ that can easily be parameterized using Gaussian assumptions. Learning and inference then can be cast as E and M step of an expectation–maximization (EM) scheme \cite{dempster1977maximum}. Usually, the E or inference step focuses on determining the the conditional mean $\mu$ of the cause states, i.e. the first moment of $q(v)$. The learning, or M, step finds the optimal value of the weights parameters $\theta$ given the expectation inferred via $\mu$. Critically, E and M step can be cast as a gradient descent scheme on the same objective function, the model's free energy $F$:

\begin{equation}
F=\ln p(u ; \theta)-K L\{q(v), p(v \mid u ; \theta)\}
\end{equation}.

Predictive coding networks minimize the free energy $F$ based on prediction errors $e$ and their precision, the inverse of the error covariance $\Sigma_{e}=\operatorname{Cov}\left(e\right)$ \cite{rao1998development, friston2009predictive}.

For a single layer PCN, the generative model has the following form:

\begin{equation}
\begin{aligned}
& u=g(v, \theta)+e^{(1)} \\
& v=\eta+e^{(2)} \\
\end{aligned}
\end{equation}

where $g(v, \theta)$ is a complex and nonlinear function parameterized by the weights $\theta$ of a (deep) neural network. Given the inferred cause state $v$ (in terms of its conditional mean $\mu$), this decoder network $g(v, \theta)$ outputs a prediction of the expected sensory observation $u$. Prior assumptions on the inferred cause are given by $\eta$ and $\Sigma^{(2)}$. Often times, $\eta$ is zero and $\Sigma^{(2)}$ is the identity matrix. More complex hierarchical models are constructed by stacking intermediate hierarchical layers:

\begin{equation}
\begin{aligned}
& v^{(1)}=g\left(v^{(2)}, \theta^{(1)}\right)+e^{(1)} \\
& v^{(2)}=g\left(v^{(3)}, \theta^{(2)}\right)+e^{(2)} \\
& v^{(3)}=\ldots
\end{aligned}
\end{equation}

where the prediction of a respective higher hierarchical layer provides an empirical (i.e. learned) prior $p\left(v^{(i)} \mid v^{(i+1)} ; \theta^{(i)}\right)=N\left(g^{(i)}, \Sigma^{(i)}\right)$ on the inferred cause state distribution of the respective lower hierarchical layer \cite{friston2009predictive, friston2003learning}.

In PCNs, the state parameters are updated via the gradient of the objective function $\dot{\mu}^{(i)}=-\frac{\partial F}{\partial \mu^{(i)}}$ and the weight parameters are updated via $\dot{\theta}^{(i)}=-\frac{\partial F}{\partial \theta^{(i)}}$. More precisely, PCNs are optimized by a gradient descent on a free energy objective that depends only on the precision weighted prediction error in each hierarchical layer:

\begin{equation}
\begin{aligned}
F= & -\frac{1}{2} \xi^{(1) T} \xi^{(1)}-\frac{1}{2} \xi^{(2) T} \xi^{(2)}-\ldots \\
& -\frac{1}{2} \ln \left|\Sigma_{e}^{(1)}\right|-\frac{1}{2} \ln \left|\Sigma_{e}^{(2)}\right|-\ldots
\end{aligned}
\end{equation}

where the precision weighted prediction errors are defined as:
\begin{equation}
\begin{aligned}
& \xi^{(i)}=\Sigma^{(i)^{-1 / 2}}_e(\mu^{(i)}-g\left(\mu^{(i+1)}, \theta^{(i)}\right) ) \\
\end{aligned}
\end{equation}

In the context of PCNS, the prediction error precision $\Sigma_{e}$ plays an important role, e.g. in parameterizing attention, by modulating the gain on prediction errors \cite{friston2009predictive}. From another perspective, the prediction error precision $\Sigma_{e}$ plays the role of parameterizing an empirical prior on the activity in lower states \cite{friston2003learning}. More elaborate PCN models cover state-dependent precisions as part of their predictions, allowing them to express dynamic attentional processes \cite{feldman2010attention}. In practice however, the prediction error precision $\Sigma_{e}$ is often assumed simply be constant (e.g. the identity matrix) during learning and the corresponding terms in the free energy objective that depend only on the precision are ignored. We will follow the same simplification here, i.e. assume that the prediction error precision itself is constant and not dynamically inferred. However, we will address precision at a later stage of prediction error processing, at the stage of the back-propagated error signal that drives parameter updates in gradient-based learning. More precisely, we will focus on the inverse covariance of the gradient of the model's squared prediction error (i.e. of the free energy objective) with respect to the optimised parameter (including cause states $v$ and weights $\theta$). We suggest that this precision of of the propagated error does not parameterize (sensory) attention or empirical priors directly, but plays an important role for the efficiency, accuracy and attentional modulation of the parameter updates since it captures the uncertainty about inferred parameters given observations. 
Since the quality of state inference influences the quality of parameter learning (and vice versa) and computational resources are limited for iterative inference (especially in online learning settings and in biologically plausible implementations), fast and reliable parameter updates are crucial. 

A particularly interesting optimisation method in this context is (approximate) Natural Gradient Descent (NGD), which takes into account information about the curvature of the objective function when updating model parameters \cite{amari1997neural}. Information about this curvature is approximated with (variants of) the Fisher information matrix (FIM) of the optimised parameters. The covariance of the score, i.e. of the gradient of the log-likelihood function with respect to a parameter $p=\{\theta, v\}$ gives information about the approximate curvature of the objective function, referred to as the empirical FIM, for a batch $B$ of observations $u$ \cite{martens2020new}. The empirical FIM for weights $\theta$ is 

\begin{equation}
\begin{aligned}
I(\theta) &= \mathbb{E}_u\left[\nabla{^2}_{\theta}\log p(u|\theta)\right] \
&= \mathbb{E}_{u}\left[\nabla_{\theta}\log p(u|\theta)\nabla_{\theta}^T\log p(u|\theta)\right] \
\end{aligned}
\end{equation}

and similarly for the inferred state mean $\mu$:

\begin{equation}
\begin{aligned}
I(\mu) &= \mathbb{E}_u\left[\nabla{^2}_{\mu}\log p(u|\mu)\right] \
&= \mathbb{E}_{u}\left[\nabla_{\mu}\log p(u|\mu)\nabla_{\mu}^T\log p(u|\mu)\right] \
\end{aligned}
\end{equation}

 A nice property of the empirical FIM is that it does not require any additional information next to the gradients of the objective function, which are already computed during standard gradient descent. The diagonal elements of the empirical FIM have found widespread use in adaptive learning rate optimizers, such as RMSProp and Adam \cite{kingma2014adam}. These, however, do not resort to the off-diagonal elements of the gradient covariance, which approximate curvature. In comparison to the empirical FIM, the true FIM is based on gradients from the model's predictive distribution \cite{martens2020new, amari1997neural}. Here, we will focus on the empirical FIM. 

\subsection{ Prediction error and gradient precision in linear PCNs}

Recent work has noticed that the prediction error precision of top-down predictions of the state parameters is equivalent to the Fisher information in linear PCNs \cite{millidge2021predictivereview}:

\begin{equation}
\begin{aligned}
\mathcal{I}\left(\mathcal{F}, \mu_l\right)=\mathbb{E}\left[\frac{\partial^2}{\partial \mu_l^2} \mathcal{F}\right]
=\mathbb{E}\left[\Sigma_{(e,l)}^{-1}\right]
=\Sigma_{(e,l)}^{-1}
\end{aligned}
\end{equation}

Similarly, the Fisher information with respect to the weights in linear PCNs is equal to the prediction error precision multiplied by the covariance of input activity \cite{millidge2021predictivereview}:

\begin{equation}
\begin{aligned}
\mathcal{I}\left(\mathcal{F}, \theta_l\right) & =\mathbb{E}\left[\frac{\partial^2}{\partial \theta_l^2} \mathcal{F}\right]
=\Sigma_{(e,l)}^{-1} \mathbb{V}\left[\mu_{l+1}\right]
\end{aligned}
\end{equation}

This approach, however, does not include the propagated bottom-up prediction error from the outgoing prediction, which also induces changes in the state parameters during inference. Furthermore, the assumption of linearity does not generally hold. 

Since we here deal with gradient-based predictive coding networks in the context of automatic differentiation, we can directly compute the gradient of the objective function with respect to the parameter in question. This, however, results in relatively large gradient covariance matrices and makes it difficult to see a direct relation to the prediction error precision, a ubiquitous quantity in the predictive coding theory. 

Recent work on Natural Gradient descent in the context of linear networks and Gaussian likelihood functions (like the free energy objective treated here) has proposed a similar, but more general factorization of the Fisher information matrix \cite{bernacchia2018exact}. They showed that the NGD update for weights parameter $\theta$ can be expressed via the covariance of input activity and the covariance of the propagated prediction error:

\begin{equation}
\frac{d \theta^{(i)}}{d t} \propto (\tilde{\Lambda}_i+\lambda I)^{-1}\left\langle e_i x_{i-1}^T\right\rangle\left(\Lambda_i+\lambda I\right)^{-1}
\end{equation}

where $\left\langle e_i x_{i-1}^T\right\rangle$ is the standard gradient, $\tilde{\Lambda}$ is the covariance of input activity and $\Lambda$ is the covariance of the propagated prediction error, estimated separately for each batch. These covariance matrices are generally smaller in size than the covariance of the standard gradient itself. In this context, the gradient of the objective function with respect to the (intermediate) outputs of the models is the propagated error signal. This propagated propagated error signal generally is not identical to the prediction error signal itself. The following sections will introduce PredProp, which uses this sort of factorization to train predictive coding networks with nonlinear neural networks. In contrast to \cite{bernacchia2018exact}, PredProp uses an empirical FIM, where the propagated prediction error signal is computed simply based on the prediction error instead of the model predictive distribution.

\section{PredProp}

PredProp uses approximate curvature information during state inference and weights learning based on a factorization of the FIM into activity and output gradient covariance. This is inspired by the factorization of the true FIM in recent work on exact natural gradients in linear networks, however in the context of nonlinear PCNs and the empiricial FIM \cite{bernacchia2018exact}. Similar factorizations have been employed in the context of training large DNN models, for example in K-FAC, where the Fisher information in the nonlinear case is efficiently computed block-wise \cite{martens2015optimizing}. In the following, we will first focus on PCNs with a single, non-linearly activated, network layer in each hierarchical layer. In this case, the outgoing prediction in each PCN $i$ layer is computed by a function $\hat{\mu}^{(i-1)} = g\left(\mu^{(i)}, \theta^{(i)}\right)$ that includes a single set of weights $\theta^{(i)}$ and a nonlinear activation function. Predprop requires only a single prediction pass (i.e. the backwards prediction from the perspective of PCNs) followed by the computation of the gradients of weight parameters and the predictions (i.e. the forward pass in PCNs), which is already a requisite for standard stochastic gradient descent.

\subsection{The M-step: Weights learning}

During learning, the gradient $g'$ of the objective function $\mathcal{F}$ with respect to the weights is modulated by the covariance of the gradients of the output $\hat{\mu}^{(i-1)}$ (i.e. the outgoing prediction) and the covariance of the input $\mu^{(i)}$ (i.e. the inferred state). In PCNs with strictly single layer decoder networks, the covariance of the gradients of the output (i.e. the propagated prediction error) $\hat{\mu}^{(i-1)}$ is closely related to the covariance of the prediction error. However, we keep the general notation here, since we are also interested in generalizing to multi-layer decoder networks later on. Based on this factorization, we can formulate updates to the generative weights $w^{(i)}$ of a particular PCN layer as:

\begin{equation}
\frac{d \theta^{(i)}}{d t} \propto \left(\Sigma_{\hat{g}}+\lambda_{\hat{g}} I\right)^{-1} g^{\prime} t\left(\Sigma_\mu+\lambda_{\Sigma_\mu} I\right)^{-1}
\end{equation}

where $g'_t$ is the gradient of the objective function $\mathcal{F}$ with respect to the weight parameters at iteration $t$ and $\hat{g}_t$ is the gradient of the cost function with respect to the outgoing prediction $\hat{\mu}$, the propagated prediction error. All covariances here and in the following E step are estimated across the batch. Similar to related work on linear networks, we use two damping parameters $\lambda_{\mu}$ and $\lambda_{\hat{g}}$ multiplied with identity matrices $I$ to help stabilize the updates \cite{bernacchia2018exact}. We found $0.001$ and $0.1$ to be good starting points for $\lambda_{\mu}$ and $\lambda_{\hat{g}}$ respectively. We use a high learning rate, e.g. $\alpha_{\text{M}}=0.5$ when applying these updates to the weights.  The pseudo-code \ref{alg:predprop} includes this M-step of the PredProp optimiser. In theory, the M step could be repeated multiple times in each iteration of the EM scheme, potentially leading to more data efficient weight parameter updates \cite{neal1998view}. Here, however, we focus on a simple approach that uses a single weights update step in combination with and iterative E step for inference.

\subsubsection{Multi-layer decoder networks}
When the decoder function $\hat{\mu}^{(i-1)} = g\left(\mu^{(i)}, \theta^{(i)}\right)$ in PCN layer $i$ consists of multiple, non-linearly activated layers with parameters $\theta^{(i)}_{j}$, the factorization can be extended to cover input activity and propagated error covariance for each individual layer $j$. In this setting, the updates for parameters $\theta^{(i)}_{j}$ are computed as

\begin{equation}
\frac{d \theta^{(i)}_{j}}{d t} \propto \left(\Sigma_{\hat{g}_{j}}+\lambda_{\hat{g}} I\right)^{-1} g^{\prime}_{j} t\left(\Sigma_{\mu_{j}}+\lambda_{\Sigma_\mu} I\right)^{-1}
\end{equation}

where the gradient of weights $g^{\prime}_{j}$ in each layer $j$ is modulated by its respective input covariance $\Sigma_{\mu_j}$ and the covariance of the gradient of its respective output $\Sigma_{\hat{g}_j}$. Here, in order to keep notation simple, we denote the intermediate activations in the decoder network as $\mu_j$ and $\hat{\mu}_j$, where the $\mu = \mu_0$ is the input to the decoder network and  $\hat{\mu}=\hat{\mu}_n$ is the final prediction for a DNN with n layers. While covariances are computed for each layer, the damping parameters are shared between all layers. 

\subsection{The E-step: State inference}

Similar to the optimisation of weight parameters, we approach the optimisation of state parameters by using a combination of the activity and error gradient covariances. During inference in the E step, the resulting updates of state parameters in PCN layer $i$ are:

\begin{equation}
\frac{d \mu^{(i)}}{d t} \propto \left(\Sigma_\mu+\lambda_{\Sigma_\mu} I\right)^{-1}\left(\Sigma_{g_t}+\lambda_g I\right)^{-1} g_t
\end{equation}

where $g_t$ is the gradient of the objective function $\mathcal{F}$ with respect to the state parameter $\mu$. $\Sigma_\mu$ is the covariance of the inferred state itself, while $\Sigma_{g_t}$ is the covariance of its gradient $g_t$. In contrast to the weights updates, the activity covariance for the states directly expresses uncertainty about the optimised parameter. Again, the covariances are damped using identity matrices $I$ and damping factors $\lambda_g$ and $\lambda_{\Sigma_\mu}$. Like the respective damping factors used in the M step, these damping could be dynamically adjusted between updates. Here, however, we treat them as fixed parameters. We found $\lambda_g=0.9$ and $\lambda_{\Sigma_\mu}=0.9$ to be good starting points for rapid, yet stable updates. Like in the M step, we use a high learning rate $\alpha_{\mathrm{E}}=0.9$ to apply these updates. Pseudo-code \ref{alg:predprop} shows these state updates in the PredProp optimiser. Unlike in the M step, we make multiple E steps in each EM iteration. Instead of terminating the E step at convergence, we use a fixed amount of inference steps, e.g. 10 or 20.

\begin{algorithm}[H]
\caption{PredProp}
\label{alg:predprop}
\begin{algorithmic}[nolist]

\Require $\theta_{\text{init}}, \mu_{\text{init}}$: Initial weight and state parameters
\Require $\alpha_{\text{M}}, \alpha_{\text{E}}$: Learning rates for weights and states
\Require $\lambda_{\mu}, \lambda_{g}, \lambda_{\hat{g}}, \lambda_{\Sigma_{\mu}}$: Damping parameters

\State $\theta_t \leftarrow \theta_{\text{init}}$ \Comment{Initialize weight parameters}
\State $t \leftarrow 0$ \Comment{Initialize timestep}

\While{not converged($\theta_t$)}
\State
\State \textbf{E step (Inference):}
\State $t_{E} \leftarrow 0$ \Comment{Initialize inference timestep}
\State $\mu_{t_E} \leftarrow \mu_{\text{init}}$ \Comment{Initialize state parameters}

\While{not converged($\mu_t$)}
\State $t_{E} \leftarrow t_{E} + 1$
\State $g_t \leftarrow \nabla_{\mu} \mathcal{F}(\theta_t, \mu_{t_E})$ \Comment{Gradients w.r.t. state parameters}
\State $\Sigma_{\mu} = Cov(\mu_{t_E})$ \Comment{Covariance of state parameters}
\State $\Sigma_{g_t} = Cov(g_t)$ \Comment{Covariance of state gradient}
\State $g_t \leftarrow (\Sigma_{\mu} + \lambda_{\Sigma_{\mu}} I)^{-1} (\Sigma_{g_t} + \lambda_{g} I)^{-1} g_t$ \Comment{Weight state gradients}
\State $\mu_{t_E} \leftarrow \mu_{t_E} + \alpha_{\text{E}} \cdot g_t$ \Comment{Update state parameters}
\EndWhile
\State $\mu_t \leftarrow \mu_{t_E}$
\State
\State \textbf{M step (Learning):}
\State $t \leftarrow t + 1$
\State $g't \leftarrow \nabla_{\theta} \mathcal{F}(\theta_t, \mu_t)$ \Comment{Gradients w.r.t. weight parameters}
\State $\hat{g_t} \leftarrow \nabla_{\hat{\mu}} \mathcal{F}(\theta_t, \mu_t)$ \Comment{Gradients w.r.t. outgoing prediction}
\State $\Sigma_{\hat{g}} = Cov(\hat{g_t})$ \Comment{Covariance of gradient of prediction}
\State $g't \leftarrow (\Sigma_{\hat{g}} + \lambda_{\hat{g}} I)^{-1} g't (\Sigma_{\mu} + \lambda_{\Sigma_{\mu}} I)^{-1}$ \Comment{Weight gradients}
\State $\theta_t \leftarrow \theta_t + \alpha_{\text{M}} \cdot g'_t$ \Comment{Update weight parameters}

\EndWhile
\end{algorithmic}
\end{algorithm}

\newpage
\section{Experiments}

To empirically evaluate the proposed method, we compare PCNs trained with PredProp, SGD and the Adam optimizer on three popular datasets for supervised and unsupervised learning: MNIST, FashionMNIST and OMNIGLOT. Here, we focus on completely unsupervised inference and learning in PCNs with three hierarchical layers and dense decoder networks. We first inspect the training of PCNs where the predictions in each hierarchical layer are made using a single set of weights that are non-linearly activated. During training, we assess the quality of the learned weights in terms of the accuracy of the prediction made from the deepest hierarchical layer, after iterative inference. In contrast to the accuracy of the prediction of the lowest PCN layer, this \enquote{global} prediction is computed using a pass through the entire network. Intuitively, global predictions with high accuracy require all layers in the PCN to maintain a consistent hierarchical generative model. 

In all experiments, we train the PCNs with a fixed amount of inference steps (10 or 20) followed by a single update of the weights given the inferred posterior states. We initialise the highest state with a zero prior plus a small value $1\mathrm{e}{-5}$. The prior states in intermediate hierarchical layers are initialised with their top-down prediction. All following experiments use densely connected weights with Glorot initialization \cite{glorot2010understanding}. We did not use biases. In all presented settings, we weight the top-down error by a constant factor of $\frac{1}{10}$, which can be interpreted as a hyper-parameter, that assigns more weights to the bottom-up error in comparison to the top-down error. Inference was done with a constant learning rate of 0.9 in all baseline models. Automatic differentiation and matrix inversion was done using PyTorch \cite{paszke2019pytorch}.

\subsection{Single-layer generative networks}

We trained PCNs with three PC layers and state sizes of 64, 64 and 128 units, with increasing size towards PCN layers closer to the data. Each PC layer consist of a single set weights followed by a hyperbolic tangent (tanh) activation function. We trained for 3000 weight update steps with a batch size of 128. We ran the PredProp optimiser using $0.0001$ and $0.1$ for $\lambda_\mu$ and $\lambda_{\hat{g}}$ and $0.9$ for both $\lambda_{\Sigma_\mu}$ and $\lambda_g$. 

While PredProp was run using a single setting for all datasets, we evaluated a variety of configurations for the baseline optimisers, SGD, Adam and SGD with momentum. Learning rates for the baselines covered $[0.1, 0.01, 0.001]$, the first order momentum covered $[0.0, 0.1, 0.9]$ and second order momentum covered $[0., 0.999]$.

\begin{figure}[h]
  \centering
  \includegraphics[width=1.\textwidth]{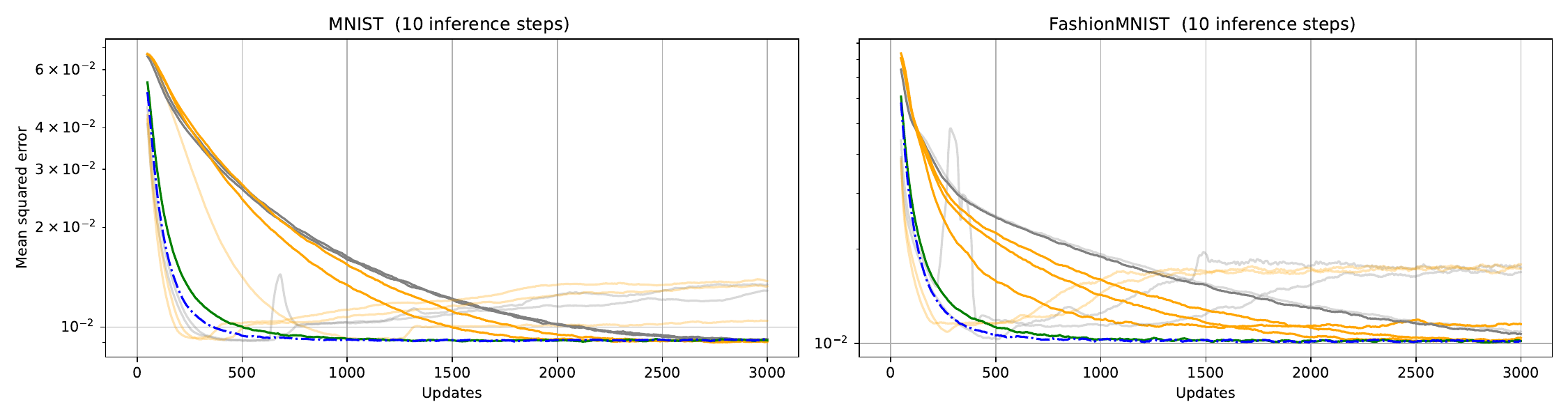}
  \includegraphics[width=1.\textwidth]{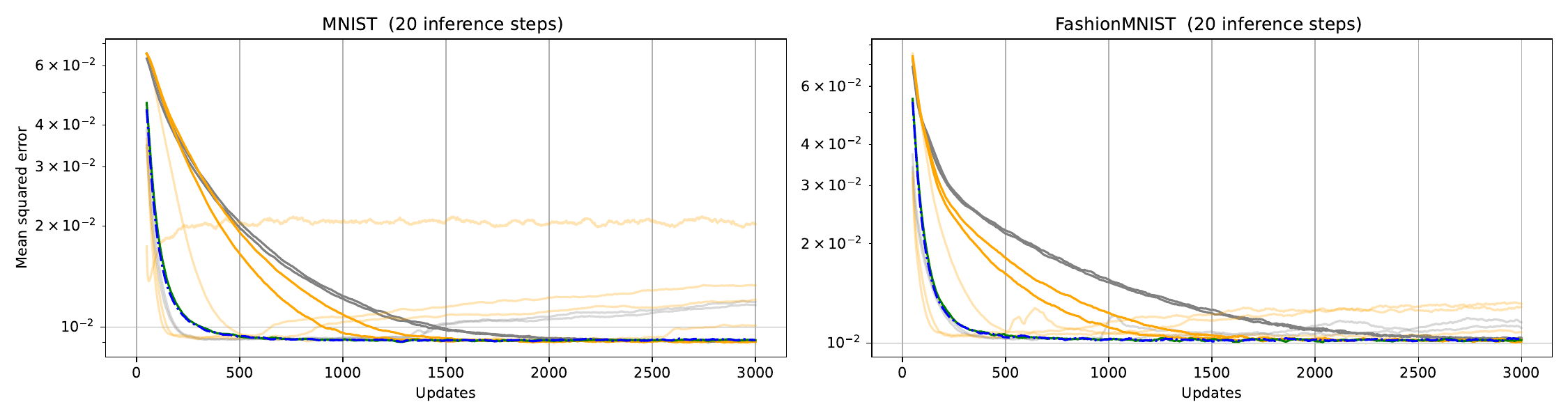}
  \caption{Training predictive coding networks with a single tanh activated layer decoding the inferred state in each hierarchical layer on MNIST and FashionMNIST. Shown are the mean squared errors during training, calculated as a running average of the previous 100 updates for PredProp and PredProp-L, a variant of PredProp that does not weight gradients during inference. For comparison, results from optimizing network weights with SGD, SGD with momentum and Adam are shown.}
  \label{fig:example}
\end{figure}

Figure \ref{fig:example} visualizes the accuracy of the global prediction of PCNs trained with PredProp, SGD and Adam. Also shown is a variant of PredProp, PredProp-L, that does not use precision weighting during iterative inference, i.e. restricts the influence of curvature information on the weights updates in the M step. We compared two different setups, one experiment with 10 iterative inference steps and another configuration with 20 inference steps. In both settings, PredProp outperforms the baseline optimization methods SGD, Adam and SGD with momentum. In all experiments with single-layer generative networks we found that including precision weighting during inference generally improves the accuracy of the model in early stages of training. In many cases, training the PCN with a baseline optimiser and high learning rates (e.g. 0.1 with SGD and momentum) allows to outperform PredProp at the start of training. These configurations, however, diverge when the model gets more accurate and more precise updates are necessary. PredProp, in contrast, also shows relatively fast convergence early on, while preventing divergence when the model is starting to get accurate.

\subsection{Multi-layer generative networks}

\begin{figure}[H]
  \centering
  \begin{subfigure}[b]{0.49\textwidth}
    \includegraphics[width=\textwidth]{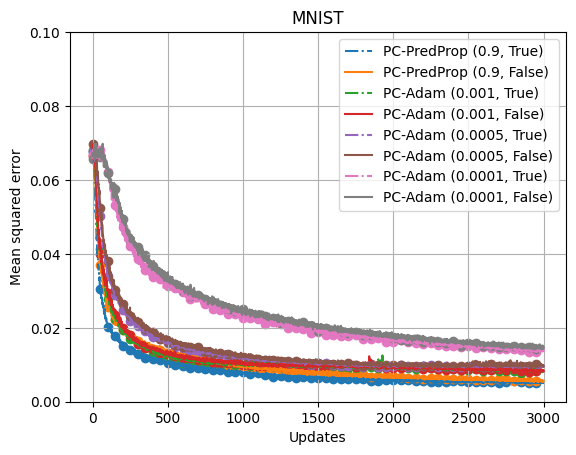}
    \label{fig:1}
  \end{subfigure}
  \begin{subfigure}[b]{0.49\textwidth}
    \includegraphics[width=\textwidth]{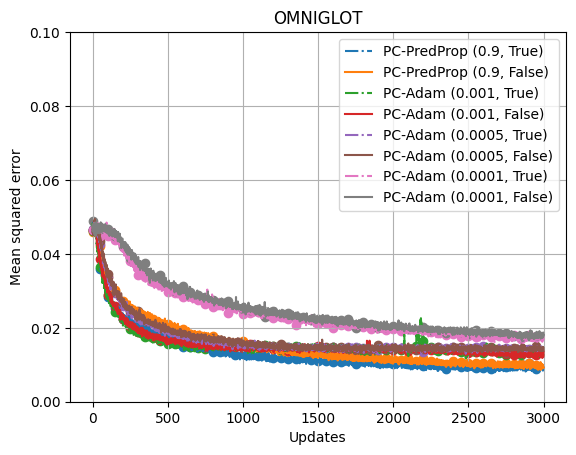}
    \label{fig:3}
  \end{subfigure}
  \caption{Training predictive coding networks with multi-layer generative networks on MNIST and OMNIGLOT. Shown are mean squared errors during training, reflecting the mean of three different runs. Circles indicate the corresponding scores from batches of the test set.  Compared are models trained using PredProp and Adam for weights optimisation. Each configuration is evaluated with (True) and without (False) PredProp's precision weighting during inference. }
  \label{fig:all}
\end{figure}

In the experiments with multi-layer generative networks, the PCNs had three layers with 64 state units each. Each PCN layer had a generative network with three layers, of which the first two were ReLU activated. The size of hidden activations in each generative networks was set to 256 units. Training was done for 3000 weight update steps with a batch size of 128 and 20 iterative inference steps. We used $0.005$ and $0.1$ for $\lambda_\mu$ and $\lambda_{\hat{g}}$ and $0.9$ for both $\lambda_{\Sigma_\mu}$ and $\lambda_g$. We compared a model trained with PredProp to baseline models trained with the Adam optimiser at learning rates covering $[0.001, 0.0005, 0.0001]$ and fixed $beta_1=0.9$ and $beta_2=0.999$ values. We trained all configurations with and without PredProp's precision weighting during iterative inference. Figure \ref{fig:all} shows train set and test set results for MNIST and OMNIGLOT. Not shown are the results on FashionMNIST, which were comparable. In the tested configurations, PredProp outperforms the baseline models trained with the Adam optimiser. Including PredProp's precision weighting lead to a noticeable increase in overall prediction accuracy in all trained configurations.

\section{Discussion and future work}

Our results indicate the usefulness of using the precision of the propagated error, and the precision of gradients more generally, when training predictive coding networks, with performance gains during iterative inference as well as learning. However, the presented experiments and evaluated baselines are relatively small scale and the employed datasets are fairly simple. Future work could investigate the role of gradient precision in more complex architectures, e.g. by including bias or using convolutional layers. Throughout or experiments, we used relatively large batch sizes and estimated covariances from scratch for each batch. Future work could improve covariance estimation, e.g. by incorporating estimations from previous batches as prior knowledge. Another interesting avenue could be to analyse the importance of precision weighting in temporal predictive coding models. The close interaction between precision weighting of the prediction errors and the propagated prediction errors yet needs to be explored in detail. Another interesting direction of research could address the influence of state-dependent prediction error precision, such as proposed in \cite{feldman2010attention} or other means of adaptive precision-based modulation in scaled-up predictive coding networks. 

\newpage
\printbibliography

@article{millidge2021predictivereview,
  title={Predictive Coding: a Theoretical and Experimental Review},
  author={Millidge, Beren and Seth, Anil and Buckley, Christopher L},
  journal={arXiv preprint arXiv:2107.12979},
  year={2021}
}

@article{friston2010generalised,
  title={Generalised filtering},
  author={Friston, Karl and Stephan, Klaas and Li, Baojuan and Daunizeau, Jean},
  journal={Mathematical Problems in Engineering},
  volume={2010},
  year={2010},
  publisher={Hindawi}
}

@article{zahid2023curvature,
  title={Curvature-Sensitive Predictive Coding with Approximate Laplace Monte Carlo},
  author={Zahid, Umais and Guo, Qinghai and Friston, Karl and Fountas, Zafeirios},
  journal={arXiv preprint arXiv:2303.04976},
  year={2023}
}

@article{feldman2010attention,
  title={Attention, uncertainty, and free-energy},
  author={Feldman, Harriet and Friston, Karl J},
  journal={Frontiers in human neuroscience},
  volume={4},
  pages={215},
  year={2010},
  publisher={Frontiers Research Foundation}
}

@inproceedings{ofner2022generalized,
  title={Generalized Predictive Coding: Bayesian Inference in Static and Dynamic models},
  author={Ofner, Andr{\'e} and Millidge, Beren and Stober, Sebastian},
  booktitle={SVRHM 2022 Workshop@ NeurIPS}
}

@article{bastos2012canonical,
  title={Canonical microcircuits for predictive coding},
  author={Bastos, Andre M and Usrey, W Martin and Adams, Rick A and Mangun, George R and Fries, Pascal and Friston, Karl J},
  journal={Neuron},
  volume={76},
  number={4},
  pages={695--711},
  year={2012},
  publisher={Elsevier}
}

@article{paszke2019pytorch,
  title={Pytorch: An imperative style, high-performance deep learning library},
  author={Paszke, Adam and Gross, Sam and Massa, Francisco and Lerer, Adam and Bradbury, James and Chanan, Gregory and Killeen, Trevor and Lin, Zeming and Gimelshein, Natalia and Antiga, Luca and others},
  journal={Advances in neural information processing systems},
  volume={32},
  year={2019}
}

@inproceedings{glorot2010understanding,
  title={Understanding the difficulty of training deep feedforward neural networks},
  author={Glorot, Xavier and Bengio, Yoshua},
  booktitle={Proceedings of the thirteenth international conference on artificial intelligence and statistics},
  pages={249--256},
  year={2010},
  organization={JMLR Workshop and Conference Proceedings}
}

@article{dempster1977maximum,
  title={Maximum likelihood from incomplete data via the EM algorithm},
  author={Dempster, Arthur P and Laird, Nan M and Rubin, Donald B},
  journal={Journal of the Royal Statistical Society: Series B (Methodological)},
  volume={39},
  number={1},
  pages={1--22},
  year={1977},
  publisher={Wiley Online Library}
}

@article{amari1997neural,
  title={Neural learning in structured parameter spaces-natural Riemannian gradient},
  author={Amari, SI},
  journal={Advances in neural information processing systems},
  pages={127--133},
  year={1997},
  publisher={MORGAN KAUFMANN PUBLISHERS}
}

@inproceedings{martens2015optimizing,
  title={Optimizing neural networks with kronecker-factored approximate curvature},
  author={Martens, James and Grosse, Roger},
  booktitle={International conference on machine learning},
  pages={2408--2417},
  year={2015},
  organization={PMLR}
}

@article{bernacchia2018exact,
  title={Exact natural gradient in deep linear networks and its application to the nonlinear case},
  author={Bernacchia, Alberto and Lengyel, M{\'a}t{\'e} and Hennequin, Guillaume},
  journal={Advances in Neural Information Processing Systems},
  volume={31},
  year={2018}
}

@article{marino2022predictive,
  title={Predictive coding, variational autoencoders, and biological connections},
  author={Marino, Joseph},
  journal={Neural Computation},
  volume={34},
  number={1},
  pages={1--44},
  year={2022},
  publisher={MIT Press}
}

@inproceedings{lin2021tractable,
  title={Tractable structured natural-gradient descent using local parameterizations},
  author={Lin, Wu and Nielsen, Frank and Emtiyaz, Khan Mohammad and Schmidt, Mark},
  booktitle={International Conference on Machine Learning},
  pages={6680--6691},
  year={2021},
  organization={PMLR}
}

@inproceedings{csenoz2018online,
  title={Online variational message passing in the hierarchical Gaussian filter},
  author={{\c{S}}en{\"o}z, {\.I}smail and De Vries, Bert},
  booktitle={2018 IEEE 28th International Workshop on Machine Learning for Signal Processing (MLSP)},
  pages={1--6},
  year={2018},
  organization={IEEE}
}

@article{friston2003learning,
  title={Learning and inference in the brain},
  author={Friston, Karl},
  journal={Neural Networks},
  volume={16},
  number={9},
  pages={1325--1352},
  year={2003},
  publisher={Elsevier}
}

@article{friston2006free,
  title={A free energy principle for the brain},
  author={Friston, Karl and Kilner, James and Harrison, Lee},
  journal={Journal of physiology-Paris},
  volume={100},
  number={1-3},
  pages={70--87},
  year={2006},
  publisher={Elsevier}
}

@article{robbins1951stochastic,
  title={A stochastic approximation method},
  author={Robbins, Herbert and Monro, Sutton},
  journal={The annals of mathematical statistics},
  pages={400--407},
  year={1951},
  publisher={JSTOR}
}

@article{neal1998view,
  title={A view of the EM algorithm that justifies incremental, sparse, and other variants},
  author={Neal, Radford M and Hinton, Geoffrey E},
  journal={Learning in graphical models},
  pages={355--368},
  year={1998},
  publisher={Springer}
}

@article{rao1999predictive,
  title={Predictive coding in the visual cortex: a functional interpretation of some extra-classical receptive-field effects},
  author={Rao, Rajesh PN and Ballard, Dana H},
  journal={Nature neuroscience},
  volume={2},
  number={1},
  pages={79--87},
  year={1999},
  publisher={Nature Publishing Group}
}

@article{parr2019neuronal,
  title={Neuronal message passing using Mean-field, Bethe, and Marginal approximations},
  author={Parr, Thomas and Markovic, Dimitrije and Kiebel, Stefan J and Friston, Karl J},
  journal={Scientific reports},
  volume={9},
  number={1},
  pages={1889},
  year={2019},
  publisher={Nature Publishing Group UK London}
}

@article{friston2009predictive,
  title={Predictive coding under the free-energy principle},
  author={Friston, Karl and Kiebel, Stefan},
  journal={Philosophical transactions of the Royal Society B: Biological sciences},
  volume={364},
  number={1521},
  pages={1211--1221},
  year={2009},
  publisher={The Royal Society London}
}

@article{martens2020new,
  title={New insights and perspectives on the natural gradient method},
  author={Martens, James},
  journal={The Journal of Machine Learning Research},
  volume={21},
  number={1},
  pages={5776--5851},
  year={2020},
  publisher={JMLRORG}
}

@article{rao1998development,
  title={Development of localized oriented receptive fields by learning a translation-invariant code for natural images},
  author={Rao, Rajesh PN and Ballard, Dana H},
  journal={Network: Computation in Neural Systems},
  volume={9},
  number={2},
  pages={219},
  year={1998},
  publisher={IOP Publishing}
}

@article{scellier2017equilibrium,
  title={Equilibrium propagation: Bridging the gap between energy-based models and backpropagation},
  author={Scellier, Benjamin and Bengio, Yoshua},
  journal={Frontiers in computational neuroscience},
  volume={11},
  pages={24},
  year={2017},
  publisher={Frontiers Media SA}
}

@article{kingma2014adam,
  title={Adam: A method for stochastic optimization},
  author={Kingma, Diederik P and Ba, Jimmy},
  journal={arXiv preprint arXiv:1412.6980},
  year={2014}
}

@article{amari1998natural,
  title={Natural gradient works efficiently in learning},
  author={Amari, Shun-Ichi},
  journal={Neural computation},
  volume={10},
  number={2},
  pages={251--276},
  year={1998},
  publisher={MIT Press}
}

\end{document}